\documentclass{article}

\usepackage{arxiv}

\usepackage[utf8]{inputenc} 
\usepackage[T1]{fontenc}    
\usepackage{hyperref}       
\usepackage{url}            
\usepackage{booktabs}       
\usepackage{amsfonts}       
\usepackage{nicefrac}       
\usepackage{microtype}      
\usepackage{lipsum}		
\usepackage{graphicx}
\usepackage[numbers]{natbib}
\usepackage{doi}
\usepackage{threeparttable}

\title{Greater than the sum of its parts: The role of minority and majority status in
collaborative problem-solving communication}

\author{{\hspace{1mm}Jacqueline G. Cavazos} \\
	School of Education\\
	University of California, Irvine \\
	\texttt{jacqueline.cavazos@uci.edu} \\
	\And
	{\hspace{1mm}Nia Nixon} \\
	School of Education\\
	University of California, Irvine\\
	\texttt{dowelln@uci.edu} \\
}

\renewcommand{\shorttitle}{\textit{arXiv} Template}

\hypersetup{
pdftitle={A template for the arxiv style},
pdfsubject={q-bio.NC, q-bio.QM},
pdfauthor={David S.~Hippocampus, Elias D.~Striatum},
pdfkeywords={First keyword, Second keyword, More},
}

\begin{document}
\maketitle

\begin{abstract}
Collaborative problem-solving (CPS) is a vital skill used both in the workplace and in educational environments. CPS is useful in tackling increasingly complex global, economic, and political issues and is considered a central 21st century skill. The increasingly connected global community presents a fruitful opportunity for creative and collaborative problem-solving interactions and solutions that involve diverse perspectives. Unfortunately, women and underrepresented minorities (URMs) often face obstacles during collaborative interactions that hinder their key participation in these problem-solving conversations. Here, we explored the communication patterns of minority and non-minority individuals working together in a CPS task. Group Communication Analysis (GCA), a temporally-sensitive computational linguistic tool, was used to examine how URM status impacts individuals' sociocognitive linguistic patterns. Results show differences across racial/ethnic groups in key sociocognitive features that indicate fruitful collaborative interactions. We also investigated how the groups' racial/ethnic composition impacts both individual and group communication patterns. In general, individuals in more demographically diverse groups displayed more productive communication behaviors than individuals who were in majority-dominated groups. We discuss the implications of individual and group diversity on communication patterns that emerge during CPS and how these patterns can impact collaborative outcomes.
\end{abstract}

\keywords{collaborative problem solving  \and socio-cognitive processes \and natural language processing \and underrepresented minorities \and text analysis \and Group Communication Analysis}

\section{Introduction}
During collaborative problem-solving (CPS) tasks, individuals work together, exchange ideas, and share information to solve a problem \citep{oecd2013pisa,oecd2017pisa}. CPS is at the center of teamwork and learning across many fields and particularly in STEM education \citep{oecd2013pisa}. The integral role of CPS in solving complex learning, workplace, and global problems makes it a key 21st century skill \citep{burrus2013identifying} and a central transversal skill for the workforce \citep{fiore2018collaborative}. In educational contexts, successful team interactions have the potential to improve not only student outcomes, but also students’ psychological lived experiences (e.g., sense of belonging, self-efficacy, and confidence) \citep{walton2007question, dasgupta2015female, niler2020solidarity, snyder2016peer,graesser2011computer,cade2014modeling}. Given the benefits of successful collaborative interactions, it is imperative that fruitful collaborative activities benefit students across diverse demographic backgrounds. To date, disparities in collaborative settings continue to impact women and traditionally underrepresented racial and ethnic minorities (URMs) (Black, Latinx, Asian American and Pacific Islanders (AAPI), Indigenous, and other minority communities) at a disproportionate rate \citep{chemers2011role}. For instance, women and URMs can often feel left out by their peers and are given limited opportunities to speak and interact with other students in face-to-face and online interactions \citep{dasgupta2014girls,varma2006confronting,ong2005body,du2015online}. These differences across minority and majority students are troublesome given that increased globalization has also increased the diversity in teams in terms of race/ethnicity, gender, and culture \citep{chun2009bridging}. Addressing these inequities can help marginalized student populations benefit from successful collaborative interactions while also allowing teams to benefits from more fruitful problem-solving discussions.

Previous research has shown that race/ethnicity can sometimes (but not always \citep{pearson2018using}) impact students outcomes and experiences during collaborative interactions \citep{cintron2019exploring,kumi2017online}. Notably, these previous findings focus on students psychological experiences or performance outcomes. However, what is less clear is how individuals communicate \textit{during} collaborative interactions and how these discussions impact student outcomes. Communication, and specifically language patterns, can help identify the important psychological processes that occur during CPS tasks \citep{fiore2010toward,graesser2018challenges,tausczik2010psychological,schneider2021collaboration}. More recently, some studies have leveraged language to explore how gender can impact the sociocognitive dynamics that occur in CPS \citep{lin2019modeling, dowell2019group, lin2020liwcs}. A few studies have explored the communication patterns that occur during collaborative interactions across different cultures \citep{kim2002cross, zhang2007impact,weinberger2013inducing} and race/ethnicity \citep{robert2018differences}. Many studies focus mostly on analyzing survey data or manually coding communication behaviors. Although these approaches provides some insight on the linguistic characteristics across race/ethnicity, communication during CPS is inherently, temporally interdependent. As such, communication patterns during collaborative settings are shaped by individuals influence on one another. This temporal interdependence between individuals can be difficult to capture via surveys. Furthermore, manual coding schemes can often be timing-consuming and constrained by available resources. Moreover, to date, no study has considered how individuals' URM status and group majority/minority composition impacts communication patterns. In the current study, we aim to investigate these differences by using computational linguistics to leverage URM and non-URM individuals' linguistic patterns. These findings can inform our understanding of the intrapersonal sociocognitive process that occur during a CPS task.

In the sections that follow, we review the background literature on disparities in CPS for underrepresented minorities, particularly across race/ethnicity. Specifically, we discuss how communication during CPS tasks can inform our understanding of the sociocognitive processes that occur during collaborative interactions. Next, we discuss what it means to be a minority and how minority status can impact racial/ethnic groups differently. Then, we review our methodological approach for exploring the impact of minority/majority status on group communication and discuss how we use an innovative natural language processing (NLP) technique, Group Communication Analysis (GCA) \citep{dowell2019group} to inform our understanding of the linguistic patterns during a CPS task across URM status. Finally, we present our results and discuss their implications in the context of understanding how CPS can be leveraged to create more equitable interactions and experiences across racial/ethnic groups. 

\section{Background}
\label{sec:headings}

\subsection{Collaborative Problem Solving}
Collaborative problem solving (CPS) involves individuals working together to exchange information, share ideas, maintain communication, and use this combined knowledge to solve a problem \citep{oecd2013pisa,oecd2017pisa}. These group processes are often separated into two domains: social and cognitive. The social domain stems from the collaborative nature of CPS. Social processes involve the social interactions between individuals including their communication and the dialogue exchanged while working through a problem. The cognitive domain stems from the problem-solving aspect of CPS. Cognitive processes involve the acquisition of knowledge, and formation and understanding of the problem. CPS is necessary for solving highly complex problems and therefore is considered a key central 21st century skill \citep{burrus2013identifying} that has been also considered essential for educational and workplace settings \citep{fiore2010toward,pena2017pisa}. Unfortunately, despite the global need for individuals to master CPS skills, a Organization of Economic Cooperation and Development (OECD) report revealed that CPS is a skill that remains poorly underdeveloped in students \citep{pena2017pisa}. Students' CPS scores from the Programme for International Student Assessment (PISA) assessment on collaboration revealed that that less than 10 percent of students performed at the highest level of CPS skills. This deficient in CPS skills extends to college students, and the incoming workforce \citep{mcgunagle2020employability,sarkar2016graduate,griffin2015assessment}. Although employers often report that collaboration and problem-solving skills are among the top skills needed for workforce readiness \citep{mcgunagle2020employability,sarkar2016graduate} they also report that recent graduates have not mastered these skills. Furthermore, recent graduates report feeling a lack of preparation in CPS skills from their undergraduate education \citep{sarkar2016graduate}. The notable gap between the global need for effective CPS skills and the lack of students' mastery in these skills underscores the need to examine how students engage in during collaborative activities in order to support the acquisition of CPS skills.

\subsection{Collaborative Interactions: Majority and Minority Students}
One clear barrier to addressing the CPS skills gap is the disproportionate disadvantage that women and URM students (e.g., Black, Latinx, AAPI, Indigenous, and other minority communities) face in collaborative group settings. Women and URMs often do not benefit from the fruitful outcomes of collaborative teamwork to the same extent as non-URM students  \citep{dasgupta2014girls,wolfe2016teamwork, micari2021beyond}. Differences across racial/ethnic minority and majority individuals are evident across outcomes such as performance \citep{oecd2013pisa,oecd2017tech} and perceptions and experiences \citep{cintron2019exploring,kumi2017online, wolfe2016teamwork}. For example, women in engineering classes report lower perceptions of learning while working in collaborative teams than men \citep{wolfe2016teamwork}. In addition, women and URM students often report being less involved in the groups' work and also dealing with more overbearing and dominate teammates than non-URM individuals. URM students also differ in their perceptions of CPS activities compared to non-URM students \citep{cintron2019exploring,kumi2017online}. Cintron et al., 2019, explored self-reported perceptions of undergraduate students URM and non-URM students across three computer science courses. Students were asked to self-report their perceptions on instructor support, collaborative learning, and motivation for their class. URM students reported lower levels of professor support as well as less positive perceptions of their collaborative learning activities than their non-URM peers. In addition, URM students found their class to be more relevant than non-URM students. Minority students have the potential to benefit from online collaborative learning, but barriers such as a lack of multicultural inclusion and a sense of marginalization, still do not allow for equitable experiences in collaborative groups across racial/ethnic lines \citep{kumi2017online}. Exploring how minority and majority individuals perceive collaborative interactions provides valuable information on their experiences in collaborative settings. However, what remains unclear is what interactions occur during collaborative activities that can lead to these disparities for different racial/ethnic groups. 

Minority and majority students can have different experiences and benefit differently from collaborative activities. However, another factor that can influence collaborative teams is the demographic composition of the group \citep{mitchell2019racial,li1999effects,zhang2007impact}. The literature provides strong support for the complex effects of team diversity on group outcomes \citep{kirkman2004impact,stahl2010unraveling}. For example, increased team heterogeneity has been shown to increase innovative solutions and satisfaction \citep{stahl2010unraveling}, but also reduce team empowerment and increase conflict \citep{kirkman2004impact,stahl2010unraveling}. The intricate role of team diversity on group outcomes is further complicated by individuals' own minority and majority status. Previous research suggests that minority individuals tend to participate less when they are a part of a majority White group compared to when they are a part of mostly minority group \citep{li1999effects}. Minority students also tend emerged as leaders in majority-dominated groups only when they self-report high levels of learning goal orientation, or the desire to learn or master skills \citep{mitchell2019racial}. Notably, computer-mediated communication (CMC) has been shown to mitigate some of the effects of group racial/ethnic diversity \citep{zhang2007impact,giambatista2010diversity,robert2018differences}. Robert et al., (2018) examined the impact of face-to-face interactions versus text-based communication on gender and race-diverse teams. Race-diverse teams that communicated via text-based interactions were found to have higher levels of knowledge sharing and information integration compared to racially diverse teams who had face-to-face interactions \citep{robert2018differences}. Text-based communications benefit from less salient demographic cues that may be more apparent in face-to-face interactions. Although these findings provide important information on the impact of racial/ethnic diversity in teams and minority versus majority status, most are limited to self-reported survey measures of team dynamics. Furthermore, previous research has primarily focused on racial/ethnic diversity rather than minority vs majority group status. Different cultures and countries may define minority and majority status differently. Individual and therefore group overall majority/minority status may have nuanced but important impacts on teams' communication behaviors.

Disentangling the complexities of how individuals and teams communicate during team CPS interactions, and more importantly, how these differences may differ across student's minority or majority status, can be a promising start to address inequalities in collaborative interactions. In addition, gaining insight into the key linguistic differences across URM and non-URM status individuals and groups can help shape interventions and specialized support strategies catered to specific groups.

\subsection{Defining an underrepresented minority}
One notable challenge in examining communication differences across different minority and majority groups, is identifying what it means to be an URM. According to the National Science Foundation \citep{nsf2014stats}, URM individuals are those who identify as ``Women, persons with disabilities, and three racial and ethnic groups—Blacks, Hispanics, and American Indians or Alaska Natives.'' By this account, all other racial/ethnic groups such as White, Asian, Asian-American and international students are considered non-URM students. This categorization is most evident in STEM fields in which Asian and White students are typically over-represented groups while URM are underrepresented \citep{national2019women}. To date, the studies that have explored the collaboration process across URM status often include Asian and White males and females \citep{cintron2019exploring} as part of non-URM students. However, recent research has highlighted the unique position of Asian and Asian-American students in society compared to other minority groups \citep{yip2021rendered,djoan2019trends,kiang2017moving}. Asian and Asian-American individuals are not often considered part of the initiatives designed to support minority groups. Recently, the American Psychologist made a call for a special issue that focused on the unique position of Asian and Asian-Americans in our society. Specifically the ``model minority'' or potentially marginalized minority \citep{yip2021rendered}. Recent reports suggest that since 1992, less than 1\% of all National Institutes of Health (NIH) funding has supported projects that are focused on Asian and Asian-American individuals \citep{djoan2019trends}. This limited funding for Asian individuals is even more troublesome considering that Asian and Asian Americans are one of the fastest group racial/ethnic groups in the United States \citep{yip2021rendered}. In addition, previous research suggests that Asian and Asian Americans can face invisible barriers towards identifying as a ``majority'' group in U.S society \citep{kiang2017moving,wing2007beyond}. Although large representation of Asian individuals in STEM fields has been well documented, their unique role in society as ``model-minorities'' sheds an important light on how Asian individuals are impacted during group interactions \citep{kiang2017moving,wing2007beyond}. Furthermore, this invisible, but often reported, boundary suggests that perhaps Asian students, and their experiences in collaborative group settings, may be different than other non-URM students. Here we aim to explore this distinction by including Asian and Asian American students as their own group and therefore not grouped under the non-URM category. As such, we explored how minority/majority status impacts Asian, URM, and non-URM students sociocognitive linguistic patterns in a CPS task. 

\subsection{Communication and Collaborative Problem Solving}
\label{sec:others}
Previous research has demonstrated the important role of communication and language in collaborative group interactions \citep{fiore2010toward,graesser2018challenges}. In fact, communication between teammates is one of the main differences between individual problem-solving and collaborative problem solving \citep{graesser2018challenges,graesser2018advancing}. Communication is a central aspect of CPS not only because it is one of the key differences between CPS and individual problem-solving \citep{fiore2010toward,graesser2018challenges}, but also because important psychological processes during CPS tasks can be identified through communication and language \citep{lin2019modeling,dowell2019promoting}. The increased use of massive open online courses (MOOCs), computer-supported collaborative communication (CSCC), and large-scale online discussion posts have provided an influx of available student data that has enabled researchers to explore students' team communication patterns at scale. Specifically, language has been used to understand the communication patterns that occur during team interactions and how these interactions can impact psychological and behavioral outcomes \citep{dowell2020exploring,dowell2019promoting,lin2019modeling,dowell2021scip}. Communication patterns during team discussions have been used to determine how linguistic characteristics and team/individual outcomes differ across a variety of contexts such as gender \citep{prinsen2007gender,sullivan2015exploring,dowell2019promoting,lin2019modeling} and race/ethnicity \citep{kim2002cross,robert2018differences}. For example, researchers have found gender differences in linguistic patterns during CPS tasks, such as differences in participation levels in online collaborative settings \citep{prinsen2007gender,sullivan2015exploring,lin2019modeling,schneider2021collaboration,dowell2021s,joksimovic2019linguistic,joksimovic2018exploring}. 

Beyond traditional surface-level measures such as participation, previous research has leveraged natural language processing (NLP) techniques to understand the social and cognitive process that occur during collaborative group interactions. For example, Lin et al., (2019) explored gender communication pattern differences using Linguistic Inquiry Word Count (LIWC) \citep{pennebaker2015development,tausczik2010psychological}, a dictionary-based analysis tool used to infer psychological process from text-based data. The authors found that women display more intrapersonal linguistic characteristics that signal more effective communication such as eliciting responses from their peers than males \citep{lin2019modeling}. More recently, Dowell et al., 2019 expand on these findings using Group Communication Analysis (GCA), a novel multi-party communication analysis that is uses semantic analysis to obtain measures of intra-and interpersonal communication behaviors. As such GCA, produces six measures: participation, internal cohesion, social impact, responsivity, newness, and communication density (More details on this can be found in section 4.2). In a similar fashion, previous research has leveraged GCA to demonstrate that both females and males tend to have more fruitful socoiocognitive language patterns when part of a female majority group in an online CPS task \citep{dowell2019promoting}. We explore racial/ethnic composition as a variable of interest here, but GCA can also be used to analyze text across other categories (gender, culture, personalities, disciplines etc). Although there have been several studies that have explored the role of gender in communication patterns, less research has been focused on exploring differences across racial/ethnic lines, including differences for URM groups \citep{robert2018differences}. Exploring how minority and majority students communication can help expand our understanding of how race/ethnicity impacts students experiences during collaborative activities. Importantly, exploring communication patterns can inform our understanding of how minority and majority students engage in collaborative teams beyond traditional survey measures \citep{pearson2018using}.

\section{Current Study}
Diverse collaborative environments have the potential to benefit both individual and group outcomes. However, in order to maximize these benefits, it is imperative that all team members (regardless of demographic background) benefit from the interaction. Previous research has used group discourse and communication to find that gender can impact the sociocognitive processes that take place during online reflective posts \citep{lin2019modeling} and CPS tasks \citep{lin2020liwcs,dowell2019promoting,dowell2020exploring}. Notably these gender differences in collaborative environments have been found even when teammates have no visible cues to indicate their teammates gender. Similarly, other forms of demographic composition (majority vs minority status) have also been shown to impact team dynamics \citep{robert2018differences,li1999effects}. Here, we examine the impact of undergraduate students linguistic patterns during a CPS task for URM, non-URM, and Asian undergraduate students. In addition, we explore the impact of groups racial/ethnic demographic composition on both individual and group communication patterns. Five group demographic compositions are analyzed: Asian Majority, URM Majority, non-URM Majority, Ethnic Parity, and Mixed Ethnicity. 
We consider not only the importance of minority/majority status in a communication behaviors from collaborative interactions, but also the delineation of Asian and Asian American students as separate from non-URM students. Furthermore, we leverage deep NLP techniques to capture the fine-grained and temporally-dependent linguistic patterns that emerge in diverse collaborative groups. Specifically, GCA helps to identify the inter-and intrapersonal communication patterns during team collaboration. The insights gained from this study have the potential to inform our understanding of how we can support URM students in CPS tasks in order to achieve more equitable learning outcomes across racial/ethnic lines. As such, we aimed to address the following questions:

\begin{quote}
    RQ1: Are there race/ethnic differences (based on URM status) in socio-cognitive linguistic features displayed during an online CPS task?
\end{quote}%
\begin{quote}
    RQ2: Does group composition impact individual students’ socio-cognitive linguistic features displayed during an online CPS task?
\end{quote}%
\begin{quote}
    RQ3: Does group composition impact the group's socio-cognitive linguistic features displayed during an online CPS task?
\end{quote}%

\section{Methods}
\subsection{Participants}
We sourced data from a large university database aimed at exploring the undergraduate experience and cognitive, psychological and academic performance outcomes. \cite{arum2021framework}. A total of 134 groups, 508 undergraduate students completed an online CPS hidden profile task. Participants were randomly assigned to a group with other students. Groups consisted of mostly four individuals and a few groups of three or five participants. In order to define group composition the same across all groups, only groups of four were used in the final analysis of 102 groups and 408 individuals. The final dataset consisted of URM students (\textit{N} = 150), Asian students (\textit{N} = 178) and non-URM students (\textit{N} = 80). URM students included, Black (\textit{N} = 7), Hispanic (\textit{N} = 133), Pacific Islander (\textit{N}= 1), and Mixed-Ethnicity (\textit{N} = 9) students. Asian students included Asian and Asian American (\textit{N} = 178) students. Non-URM students included White/non-Hispanic (\textit{N} = 53) and International (\textit{N} = 27) students. Five racial/ethnic group compositions emerged from the data: Asian Majority (\textit{N} = 22), URM Majority (\textit{N} = 16), non-URM Majority (\textit{N} = 4),  Ethnic Parity (\textit{N} = 23), Mixed Ethnicity (\textit{N} = 37). Asian Majority, URM Majority, and non-URM Majority were each defined as groups that had three or more members of each respective group. For example, Asian Majority groups included at least three members that identified as Asian. URM Majority groups included at least three URM members, and non-URM Majority included at least three non-URM members. Ethnic Parity groups were defined as groups with equal distribution of two groups (e.g., two URM/two non-URM, two Asian/two URM or two Asian/two non-URM). Finally, all other groups were categorized as Mixed Ethnicity groups. For example, Mixed Ethnicity groups could include one URM student, on Asian student, and two non-URM students or any other 1:1:2 student ratio.

\subsection{Quantifying Conversations: Group Communication Analysis}

To explore the sociocognitive communication patterns of URM, Asian, and non-URM individuals during a CPS task, we employed Group Communication Analysis (GCA) \citep{dowell2019group}. GCA is a computational linguistic analysis that uses text-based data from multi-party interactions to examine the fine-grained time-sensitive discourse that occurs during collaborative activities. GCA draws on Latent Semantic Analysis (LSA) and cross-correlational and autocorrelational analysis used in time-series analysis to extract the semantic relationship between individuals contributions in the group discourse. The goal of GCA is to capture the structural dynamic of the conversation (when individuals contribute), but also the semantic-relatedness and cohesion that takes place in conversation (what individuals contributed). To capture these sociocognitive processes, GCA leverages both the semantic structures and temporally-interdependent conversations of online group communication. GCA results in six measures: \textit{participation}, \textit{internal cohesion}, \textit{social impact}, \textit{overall responsivity}, \textit{newness} and \textit{communication density}. \textit{Participation} measures the number of contributions of each member of the group compared to the contributions of all other members in the group. Compared to the other five measures, participation is the only measure that does not rely on semantic-based metrics, and instead relies on more traditional surface-level measures of mean individual contributions. \textit{Internal cohesion}, measures how semantically similiar one individual's contribution is to their previous contributions during the task. As such, internal cohesion was designed to be a reflection of individuals self-regulation and self-monitoring tendencies, two key skills necessary for successful group outcomes \cite{chan2012co}. \textit{Social Impact}, is a measure of how much one individuals' contributions trigger follow-up responses from their teammates. This measure draws on the literature of co-regulation and its role in collaborative interactions \cite{azevedo2004can}. \textit{Overall Responsivity}, is a measure of how much one individual responded to their teammates contributions. This measure focuses on how much an individual takes up their teammates contribution and uses it to move the conversation along. As such, this measure draws on the literature of monitoring, regulating, and integrating teammates information \cite{dowell2019group}. \textit{Newness}, is the proportion of new information that the individual provides to the conversation. The sixth measure is Communication Density, the amount of semantically meaningful information related to the number of words used to deliver that information. Communication Density was not analyzed for the current study. At the time of the analysis, this version of GCA was being moved to an API and Communication Density was not finalized at the time. However, given the short nature of this CPS task, lacking this measure did not have crucial implications related to the current CPS environment. Combined, these GCA measures provides important information about the intrapersonal and interpersonal communication that occurs during online collaborative activities.

This novel computational linguistics approach has been tested for reliability and validity across three large datasets \cite{dowell2019group}. GCA attempts to tackle some aspects of multiparty communication, and although it is not designed to be exhaustive, its novel ability to extract key sociocognitive processes in collaborative processes make it a useful tool for taking a deeper look at communication patterns in teams. Previous research has shown that GCA can be used to capture the gender differences in communication behaviors during collaborative group settings \citep{dowell2019promoting}. For example, in a study looking at female-majority, male-majority, and gender-parity groups, females in female-majority groups engaged in more fruitful inter and intrapersonal behaviors (i.e., greater internal cohesion, social impact, overall responsivity) than females in either of the other two groups. In addition, the authors found that males engaged in more socially engaged patterns (e.g., social impact and overall responsivity) when there were more females in the group \citep{dowell2019promoting}. GCA has also been used to understand emergent roles (e.g.,Lurkers, Followers) in CPS tasks \citep{dowell2020exploring} and also in combination with other computational techniques such as Social Network Analysis (SNA) \citep{dowell2021scip}. However, to date no study has leveraged GCA to explore racial/ethnic differences in collaborative environments.

\section{Procedure}
A CPS hidden profile task was administered via the ETS Platform for Collaborative Assessment and Learning (ECPAL) platform. Participants completed the task on separate computers and were randomly assigned to work virtually in a group with other students in the same classroom. The task was divided into two general parts. First, participants were asked to rank a set of options from a given category (e.g., best job candidate). To help participants in their ranking decisions, they were are given a set of characteristics (e.g., list of pros and cons) for each option. In each group, individuals were given the same information (shared information) and also information that only they could see (unique information). After reviewing the characteristics, participants are asked to rank the three options from most favorable to the least favorable. Participants were given ten minutes to review their list and rank the options. Next, participants were asked to discuss the task and the ranking options with their teammates via a text-based chat. In order to reach the most optional ranking decision, individuals had to combine their unique information. The goal was for teams to discuss not only their shared information but their unique information as well. Successful ranking was achieved when individuals were able to effectively pool their information together. After discussing the ranking options with their teammates for 20 minutes, participants were once again asked to provide individual rankings for their given scenario. Participants could enter any ranking order and they did not have to match their teammates rankings. 

\section{Results}
\paragraph{URM Status impact on individual sociocognitive linguistic features.}
To examine the role of a student's URM status on the individual sociocognitive linguistic features (RQ1), we performed a mixed effects model for each of the five GCA measures. GCA scores were included in each model separately as the dependent variable, and student URM status was included as the independent variable. Given that groups were given different task types (e.g., categories), we include task type as a random effect. The null model contained only the random effects of task type. For all models Asian status was used as the reference group. Results from the likelihood ratio tests reveal that the full model for Participation yielded a significantly better fit than the null model, ${\chi}^2$(2) = 7.936, \textit{p} = .019, $R^2$m = .02, $R^2$c = .02. The full model for Internal Cohesion was marginally significantly better fit than the null model, ${\chi}^2$(2) = 5.98, \textit{p} = .05, $R^2$m = .01, $R^2$c = .02. For  Social Impact, ${\chi}^2$(2) = 1.17, \textit{p} = .55, $R^2$m = .00, $R^2$c = .03, Overall Responsivity, ${\chi}^2$(2) = 5.46, \textit{p} = .07, $R^2$m = .01, $R^2$c = .03, and Newness, ${\chi}^2$(2) = 4.23, \textit{p} = .12, $R^2$m = .01, $R^2$c = .01, we found that the full model was not significantly better fit than the null model. Estimates ($\beta$) and 95\% confidence intervals (\textit{CI}) for all models are found in Table 1. Both URM (\textit{M} = -0.12, \textit{SD} = 1.01) and non-URM students (\textit{M} = -0.12, \textit{SD} = 1.01) had significantly lower levels of participation than Asian students (\textit{M} = 0.15, \textit{SD} = 0.98) (Figure \ref{fig1}). In addition, non-URM individuals (\textit{M} = 0.22, \textit{SD} = 1.04) had slightly higher levels of internal cohesion than Asian individuals \textit{M} = -0.08, \textit{SD} = 0.94), although this was only marginally significant.

\begin{figure}
  \centering
  \includegraphics[width=\linewidth]{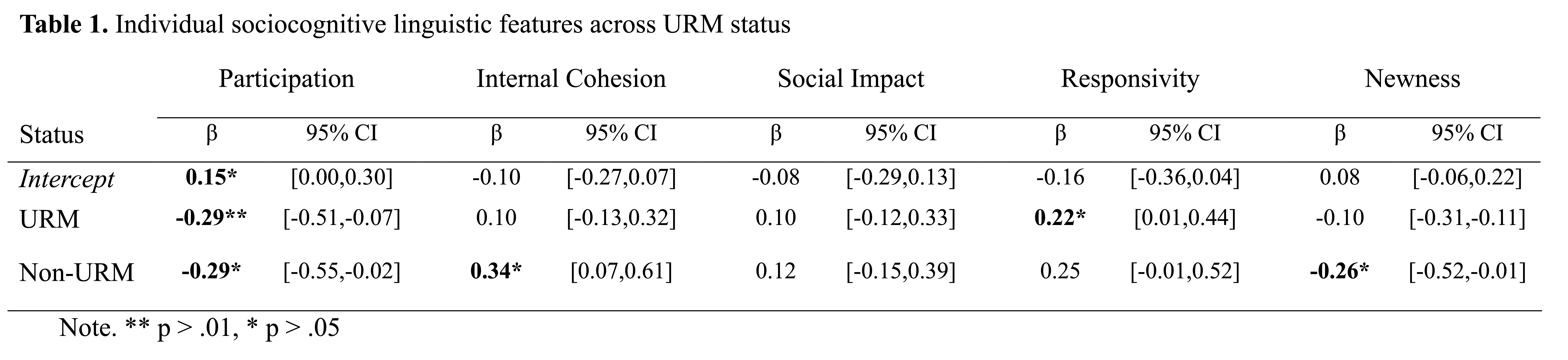}
  \label{fig1table1}
\end{figure} 

\begin{figure}
  \centering
  \includegraphics[width=\linewidth]{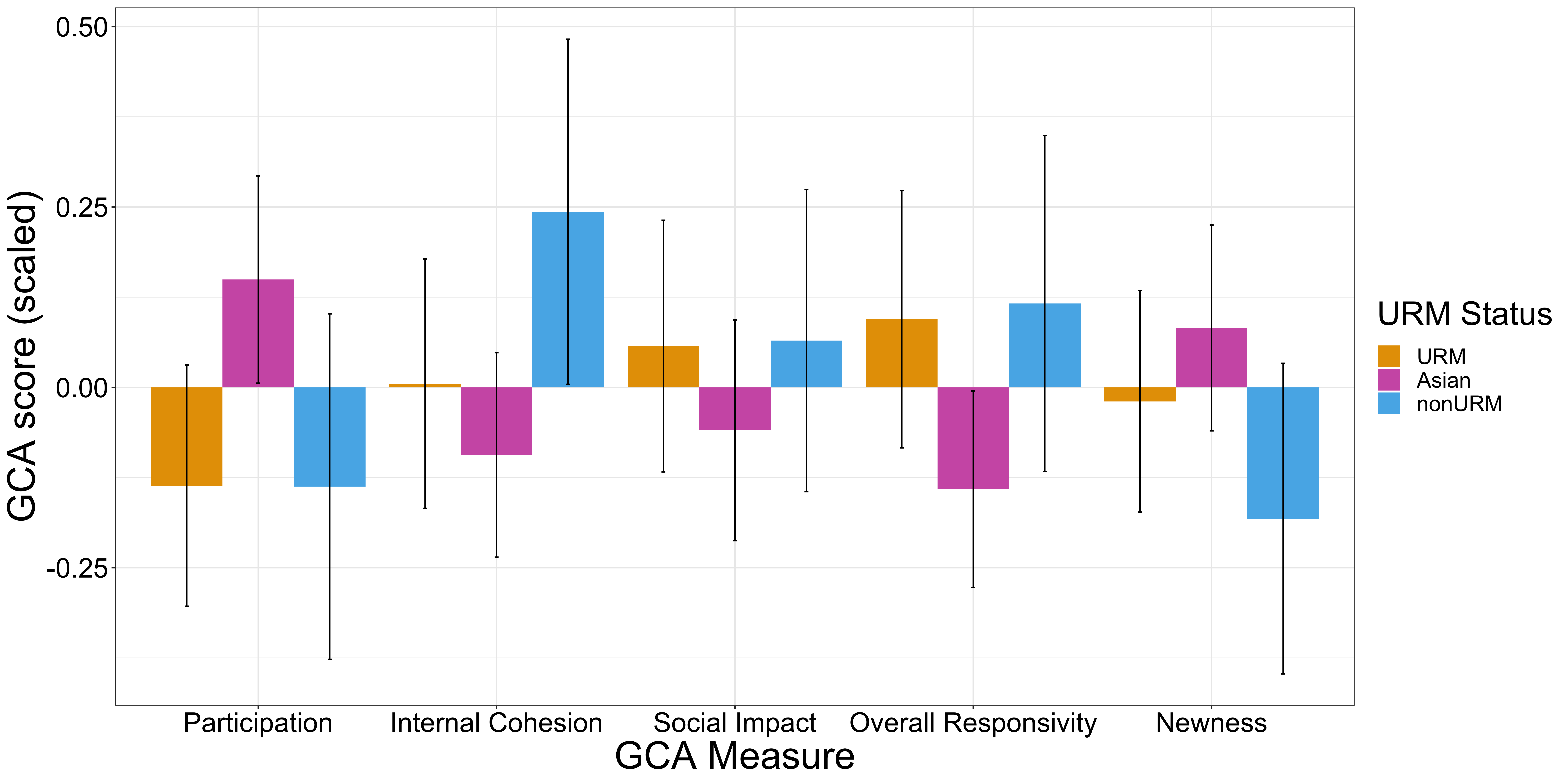}
  \caption{Individuals' mean GCA scores across different student URM status groups. Error bars depict 95\% CIs.}
  \label{fig1}
\end{figure} 

\paragraph{Group composition impact on individual sociocognitive features.}
For RQ2, we examined the role of group composition on the individual sociocognitive linguistic features. As with RQ1, five mixed effects models for each GCA measure were performed and each measure was included separately as the dependent variable. Group racial/ethnic composition was included as the independent variable. The null model contained only the random effects of task type. For all models, Asian Majority was used as the reference group. Likelihood ratio tests revealed that the full model for Participation did not yield a significantly better fit than the null model, ${\chi}^2$(4) = 0.01, \textit{p} = 1, $R^2$m = 0, $R^2$c = 0. The full model for Internal Cohesion was a significantly better fit than the null model, ${\chi}^2$(4) = 13.403, \textit{p} = .009, $R^2$m = .03, $R^2$c = .03. For Social Impact, the full model was a significantly better fit than the null model, ${\chi}^2$(4) = 13.53, \textit{p} = .009, $R^2$m = .03, $R^2$c = .05. The model for Overall Responsivity was also significantly better fit than the null model, ${\chi}^2$(4) = 13.42, \textit{p} = .009, $R^2$m = .03, $R^2$c = .05. Finally, for Newness, ${\chi}^2$(4) = 8.72, \textit{p} = .07, $R^2$m = .02, $R^2$c = .02, we found that the full model was not significantly better fit than the null model. Estimates ($\beta$) and 95\% confidence intervals (\textit{CI}) in Table 2 demonstrate that group composition impacted three of the five GCA measures. Individuals who were a part of Asian Majority groups (\textit{M} = -0.29, \textit{SD} = 0.81) had significantly lower levels of internal cohesion compared to individuals who were part of non-URM Majority (\textit{M} = 0.45, \textit{SD} = 1.21), Ethnic Parity (\textit{M} = 0.03, \textit{SD} = 1.05), and Mixed Ethnicity groups (\textit{M} = 0.15, \textit{SD} = 1.07) (Figure ~\ref{fig2}). Similarly, individuals who were part of Asian Majority groups (\textit{M} = -0.31, \textit{SD} = 0.73)  had lower levels of Social Impact compared to individuals who were a part of URM Majority (\textit{M} = 0.09, \textit{SD} = 1.07), non-URM Majority (\textit{M} = 0.50, \textit{SD} = 1.30), and Mixed Ethnicity (\textit{M} = 0.12, \textit{SD} = 1.08) groups. Finally, individuals in Asian majority groups (\textit{M} = -0.32, \textit{SD} = 0.72) had lower levels of overall responsivity compared to all other group compositions, URM Majority (\textit{M} = 0.10, \textit{SD} = 1.09), non-URM Majority (\textit{M} = 0.48, \textit{SD} = 1.10), Ethnic Parity (\textit{M} = -0.01, \textit{SD} = 1.14), Mixed Ethnicity (\textit{M} = 0.09, \textit{SD} = 1.03).

\begin{figure}
  \centering
  \includegraphics[width=\linewidth]{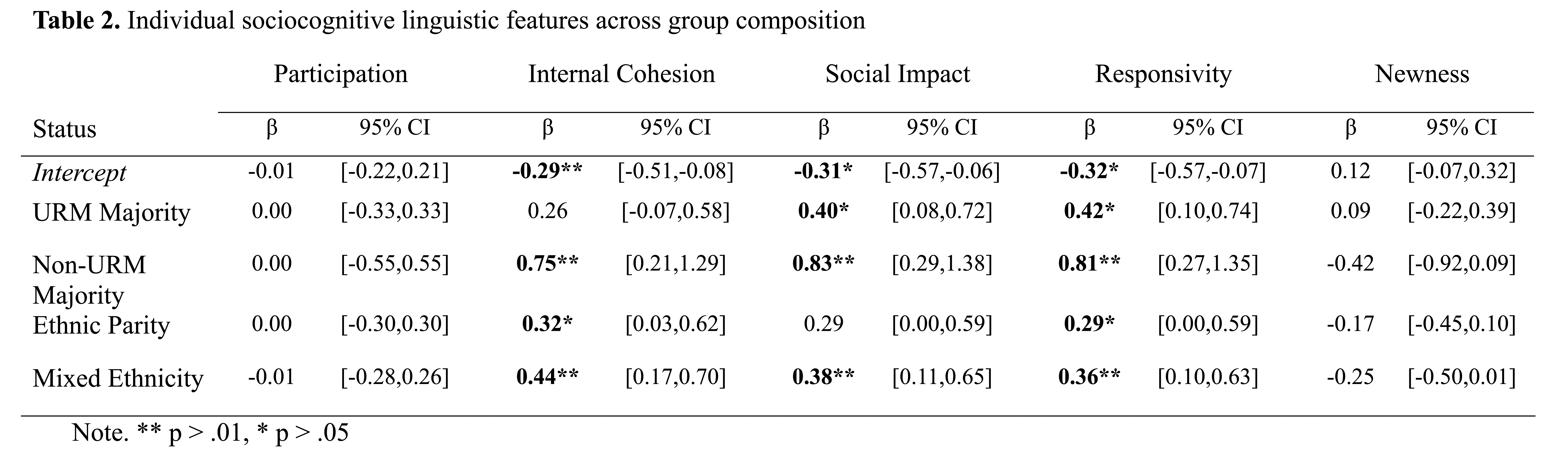}
  \label{figtable2}
\end{figure}

\begin{figure}
  \centering
  \includegraphics[width=\linewidth]{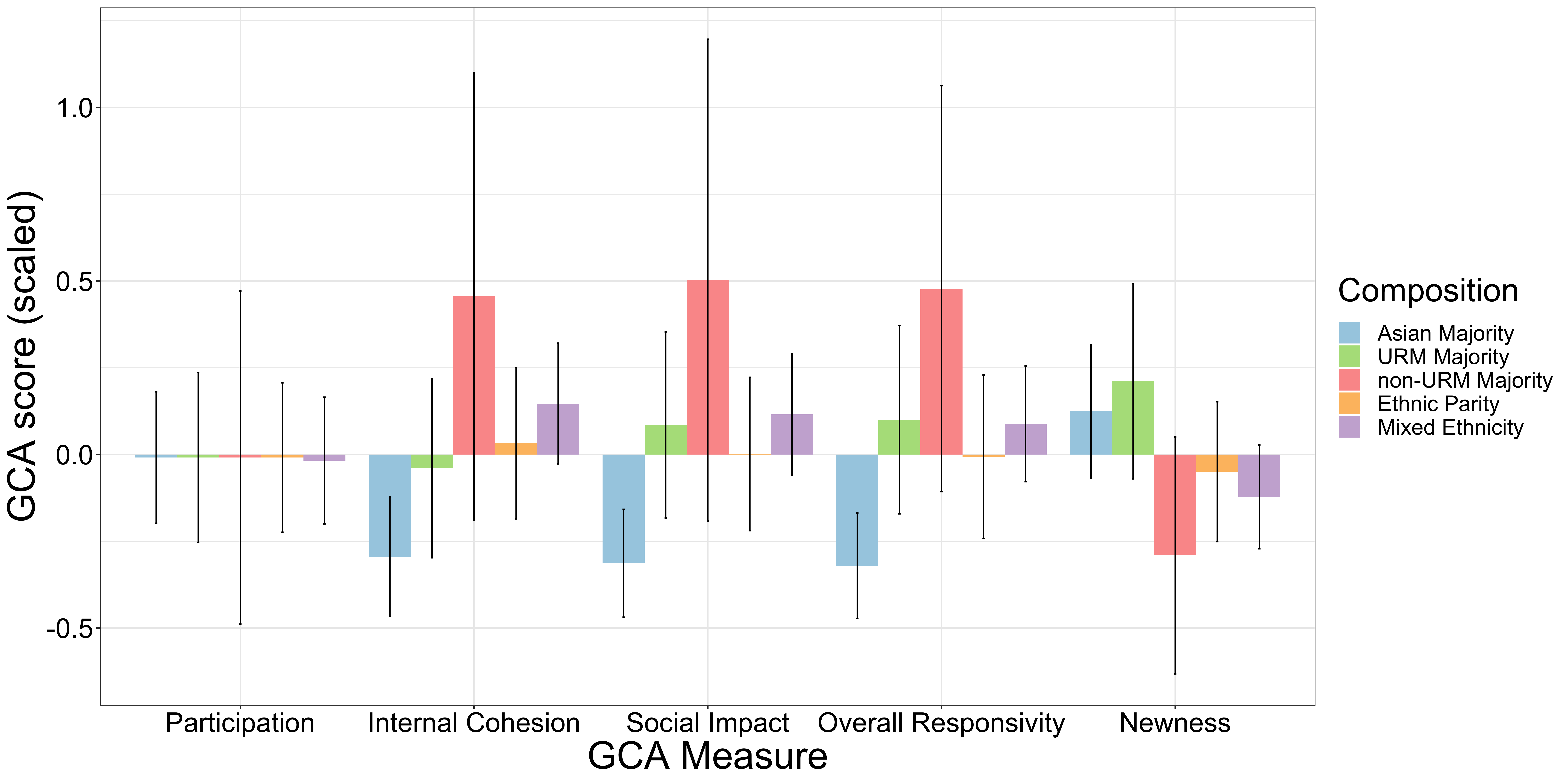}
  \caption{Individuals' mean GCA scores across group compositions. Error bars depict 95\% CIs. }
  \label{fig2}
\end{figure}

\paragraph{Group composition impact on group sociocognitive features.} For RQ3, we examined the role of group composition on the individual sociocognitive linguistic features. As with RQ1, five mixed effects models for each GCA measure were performed and each measure was included separately as the dependent variable. Group racial/ethnic composition was included as the independent variable. As with the previous models, the Asian Majority group was selected as the reference group. The null model contained only the random effects of task type. Likelihood ratio tests revealed that the full model for Participation did not yield a significantly better fit than the null model, ${\chi}^2$(4) = 1.789, \textit{p} = .77, $R^2$m = .02, $R^2$c = .02. The full model for Internal Cohesion was a significantly better fit than the null model, ${\chi}^2$(4) = 9.911, \textit{p} = .041, $R^2$m = .09, $R^2$c = .09. For Social Impact, ${\chi}^2$(4) = 5.88, \textit{p} = .207, $R^2$m = .06, $R^2$c = .07, Overall Responsivity, ${\chi}^2$(4) = 5.85, \textit{p} = .210, $R^2$m = .06, $R^2$c = .06. Finally, for Newness, ${\chi}^2$(4) = 5.31, \textit{p} = .256, $R^2$m = .05, $R^2$c = .05, the full model was not significantly better fit than the null model. Estimates ($\beta$) and 95\% confidence intervals (\textit{CI}) for all models are shown in Table 3. Internal Cohesion was the only GCA measure that group composition was able to predict over and above the task type. Asian Majority groups had lower levels of Internal Cohesion (\textit{M} = -0.31, \textit{SD} = 0.57), compared to non-URM Majority groups (\textit{M} = 0.50, \textit{SD} = 0.52) and Mixed Ethnicity group (\textit{M} = 0.12, \textit{SD} = 0.84)(Figure \ref{fig3}).

\begin{figure}
  \centering
  \includegraphics[width=\linewidth]{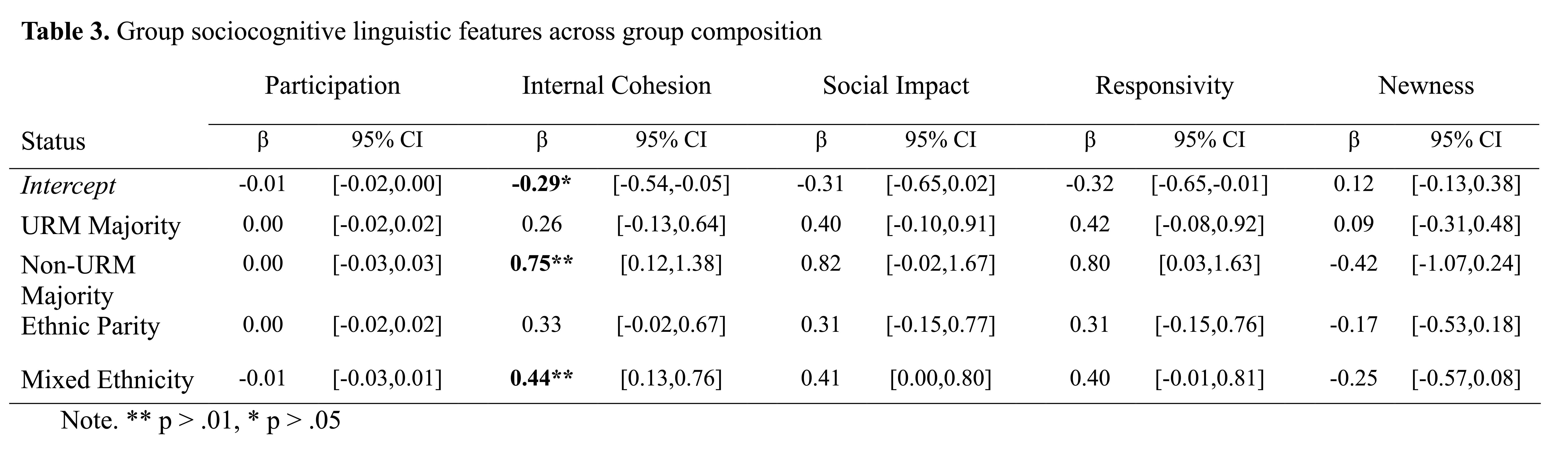}
  \label{figtable3}
\end{figure}

\begin{figure}
  \centering
  \includegraphics[width=\linewidth]{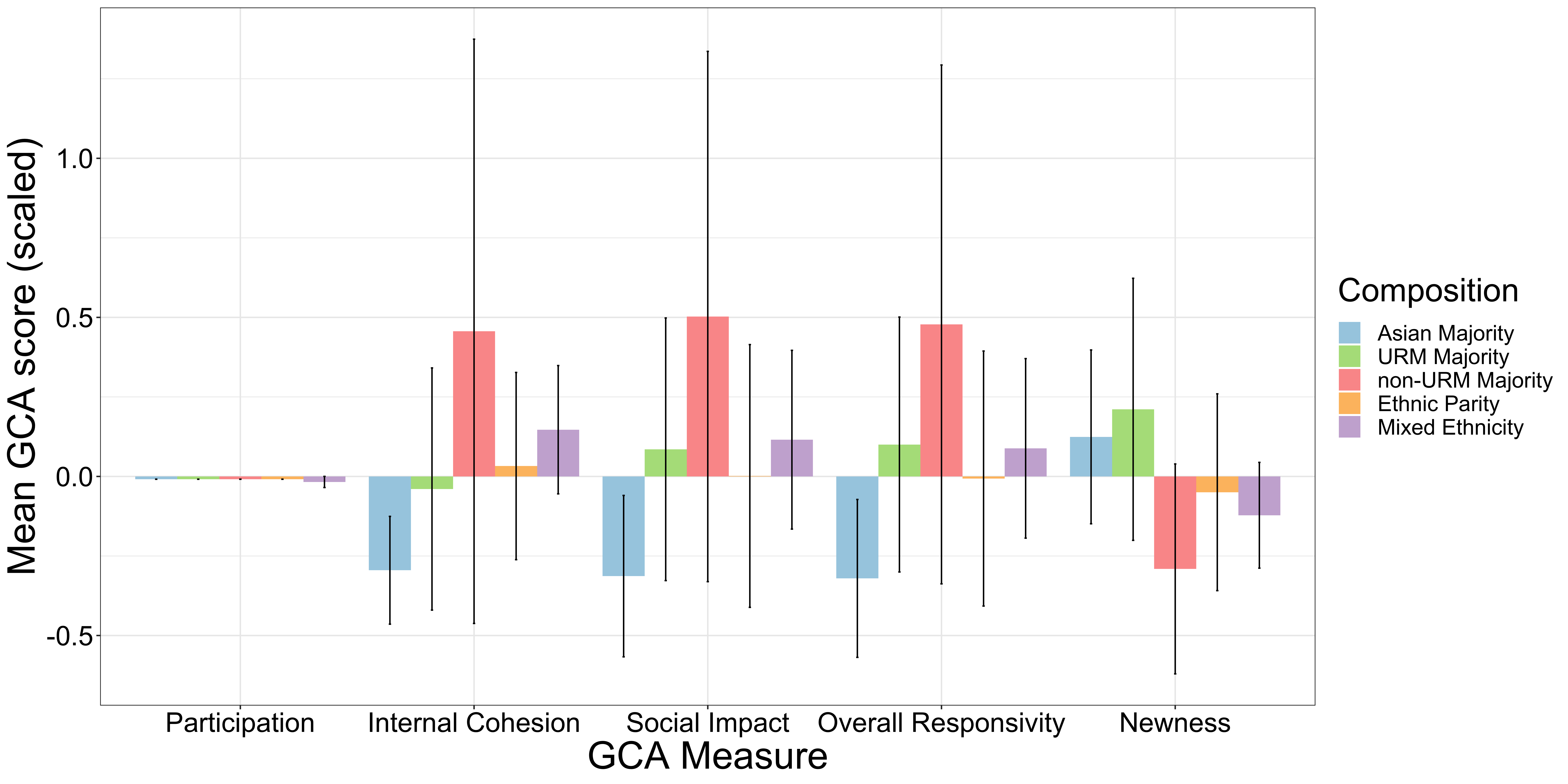}
  \caption{Group GCA scores across different group compositions.}
  \label{fig3}
\end{figure}

\section{Discussion}
The rise of globalization and international communication has increased opportunities for group collaboration among racially/ethnically diverse teams. Unfortunately, previous research demonstrates that racial/ethnic minorities do not benefit from the same collaborative experiences as individuals from traditionally majority groups \citep{dasgupta2014girls,micari2021beyond, wolfe2016teamwork}. Communication is a central aspect of group collaboration and can provide central insights into the underlying processes that occur during collaboration \citep{fiore2010toward,graesser2018challenges}. For example, previous studies have explored how communication behaviors during CPS impact students sociocognitive linguistic patterns across gender \citep{lin2019modeling,lin2020liwcs,dowell2019promoting} and cultural lines \citep{weinberger2013inducing}. However, less research has focused on how individual and group minority/majority status can impact the fundamental social and cognitive processes underlying CPS \citep{robert2018differences}. Here, we leveraged a theoretically-grounded NLP technique, GCA \citep{dowell2019group} to identify the the linguistic communication patterns of minority and majority individuals across different ethnic compositions (i.e., majority and minority groups) during a CPS task. Given the increased call to consider the struggles of Asian students as minorities \citep{micari2021beyond,djoan2019trends}, we also expanded our analysis to include Asian and Asian American students independently from non-URM students. Individuals minority/majority status impacted students' individual communication patterns (mostly participation) during CPS tasks. In addition, the demographic composition of the group influenced the individual sociocognitve linguistic patterns and to a less extent groups' sociocognitive linguistic patterns. In what follows, we discuss the implications of our findings, a general focus for future directions, and our study limitations.

Individual URM status mainly impacted Asian students' sociocognitive linguistic patterns. Asian students had the highest levels of participation which suggests that they were relatively active members in their collaborative team compared to both URM and non-URM students. Notably, Asian students' levels of participation did not align with the group they are traditionally grouped with (non-URM) nor the group they are generally compared against (URM). Asian students also had relatively lower levels of internal cohesion compared to non-URM students. Internal cohesion has been previously linked with less self-regulatory skills and greater tendency for simply echoing other teammates' views \citep{dowell2021scip,dowell2019group}. These findings suggest that compared to non-URM students, Asian students' high participation was coupled with less consistent and more surface-level contributions. In contrast, non-URM students participated less in their team interactions. However, non-URM students' contributions were more consistent (higher internal cohesion) compared to Asian students' contributions. High levels of internal cohesion have also been shown to be associated with a static mindset and inability to expand beyond ones initial thoughts \citep{dowell2021scip}. 
What is clear from these findings is that students identification as either a URM, non-URM or Asian student, did not impact individual sociocognitive linguistic patterns beyond participation. Other measures of fruitful team interactions, such as social impact and responsivity \citep{dowell2021scip,lin2019modeling,dowell2019group}, did not differ across URM, non-URM and Asian students. These findings support previous research that finds that race/ethnicity impacts team interactions, particular participation \citep{kim2002cross}. 
The limited influence of individual racial/ethnic background on students' communication patterns is even more informative when we consider that group majority/minority composition did impact individuals interpersonal and intrapersonal communication behaviors beyond participation. 

Students' communication patterns were more impacted by the demographic composition of their team, than by their own identification as an URM, non-URM or Asian individual. In this case, the group played a more crucial role in the communication behaviors of the individual than the individual members. Prior research has found a similar greater influence of the group (rather than the individual) on group outcomes \citep{kuhn2020talking,cen2016quantitative} Furthermore, our findings support previous research that shows that majority- and minority-dominated groups can impact individuals communication behaviors such as knowledge sharing and integration \citep{robert2018differences}. We expand on this knowledge by demonstrating that group demographic composition mostly influenced three individual GCA measures: internal cohesion, social impact, and overall responsivity. In general, individuals in Asian Majority groups had consistently lower levels of all three linguistic features compared to individuals in non-URM Majority and Mixed Ethnicity groups. Combined, these GCA measures have been previously been found to be markers of engaged and more fruitful collaborative interactions \citep{dowell2021scip,dowell2019group,lin2019modeling,dowell2020exploring}. For example, students with high levels of internal cohesion, social impact, and overall responsivity take on important roles within the team and display productive collaborative discussions that signal deliberate and thoughtful involvement in the task (see discussion on Drivers and Influential Actors in \citep{dowell2019group}). The potentially limited collaborative interactions by members in Asian Majority groups may have been driven by the homogeneous nature of the group (we further discuss the impact of group diversity below) \citep{wang2019team}. Our findings expand on previous findings by demonstrating that these communication differences across majority/minority groups can also lead to more engaged and productive collaborative interactions. Specifically, communications by impacting how individuals impact their teammates conversations. Not only does group composition impact and individual coherency within their own thoughts (internal cohesion), but it also impacts how they respond to others (overall responsivity) and how they prompt responses from others (social impact).

Our findings support previous findings \citep{wang2019team} that suggest that the influence of diversity in teams is complex. In the current study, Asian Majority teams were consistently the most homogeneous groups based on race. Each Asian Majority team had at least three Asian and Asian American students. In contrast, URM Majority groups had at least three URM individuals which could include either Hispanic, Black, Pacific Islander, or Mixed-Raced individuals. Non-URM Majority groups were made up of at least three non-URM individuals which included either international students and White/non-Hispanic individuals. By this account, Asian Majority groups were always represented by majority Asian students while URM and Non-URM Majority groups were not necessarily dominated by any one racial/ethnic group. As the most homogeneous group, members of Asian Majority groups displayed the lowest levels of internal cohesion, social impact, and overall responsivity. These findings suggest that the high level of homogeneity in the group may have negatively impacted the communication behaviors that typically result in more engaged CPS outcomes. One possible explanation for this finding is that groups dominated by one racial group may be less inclined to engage with their teammates in a way that can be captured by sociocognitive contributions to the team. In contrast, the non-URM Majority group with potentially the second highest level of group diversity had the highest levels of internal cohesion, social impact, and overall responsivity. Here we find that groups dominated by typically ``majority'' individuals actually produced the more engaged communication behaviors. The homogeneity of the group benefited individuals, perhaps by removing any potential for conflict or cultural/ethnic communication differences. 

Our study has a few key limitations. First, our sample study was taken from a institution that is designated both as a Hispanic-Serving Institution (HSI) and an Asian American Native American Pacific Islander-Serving Institution (AANAPISI). This designation suggests that our student sample was drawn from a population where Hispanic and Asian students may be viewed as the ``majority''. In fact, Hispanic and Asian students made up the majority of the participant pool in our study. Although it is possible that attending a minority-dominated university can have impacts on students perceptions and team interactions, we note that students' experiences as minority/majority individuals are shaped by their minority/majority status both within and outside of their educational institutions' demographic breakdown. To tease apart the potential impact of institution demographic representation, it would be beneficial for future studies to examine minority individuals experience communication behaviors when they are a part of a majority-dominated institution. Second, and related to the first limitation, our group composition sample for non-URM Majority was relatively small (\textit{N} = 4) compared to the other four group compositions. This low sample for non-URM groups may be responsible for the large degree of variability within the non-URM Majority group (large error bars). However, even with the smaller sample size we still found consistent similiarities between non-URM Majority groups and Mixed Ethnicity groups for which we had a larger sample size (\textit{N} = 37). Third, the proportion of variance explained was relatively low across all analyses performed which suggests that even though individual URM status and even more so group composition predicted some GCA measures, the proportion of variance explained by this relationship was low. One potential reason for this could be that students received only subtle cues about the demographic composition of their teammates. Notably, research suggests that computer-mediated communication can help to mitigate, and therefore reduce, some of the potential effects of racial/ethnic diversity in team activities \citep{robert2018differences}. In our study, students did not receive any audio or visual cues that revealed their teammates racial/ethnic background. Unless students revealed these details in the chat discussion, students were unaware of the demographic composition of their team. Given the design of our study were are unable to investigate how explicit cues about their teammates demographic could impact communication patterns. However, future studies could explore how/if these linguistic patterns differ when student are provided with some cues (e.g., audio, visual, explicitly disclosing) regarding their teammates demographic background. Notably, the findings from the present study still showed differences in communication patterns based on minority/majority status. This result suggests that even some indirect subtle cues based on minority/majority status were enough salient to impact their communication behaviors. To date, GCA, is the most comprehensive NLP technique that can capture the temporally-sensitive content from group communication.  Although GCA-derived measures are not exhaustive, (there are other behaviors important for CPS), the insights gleaned from GCA measures provides rich information about the teams sociocognitive processes.

\section{Conclusion}

Our study provides three central contributions to the present literature. First, we demonstrate how students' identification with majority and minority groups can impact team communication in an online collaborative team setting. Specifically, we focus on both how individual URM/non-URM status and the groups' composition of URM and non-URM individuals can play a role in the inter- and intrapersonal communication behaviors that emerge when individuals work together. These differences based on majority/minority group composition informs our understanding of how diversity, beyond gender and culture, can shape how individuals communicate in collaborative environments. Second, the delineation of Asian and Asian American students as distinct from traditional ``majority'' students adds to the increasing call to consider their struggles as a ``model'' minority group and highlights the importance of considering the experiences of various minority groups. Third, our methodological approach leveraged deeper NLP techniques in order to capture how minority/majority status impacts linguistic patterns that emerge in diverse collaborative groups. This approach allowed us to examine the complex dynamics involved in team interactions and allowed us to capture the underlying differences in communication. Differences in inter- and intrapersonal communication based on group composition were coupled with similar levels of individual participation across all groups. This finding highlights the need to explore team dynamics beyond levels of participation. The deeper level measures accessible via GCA, revealed important sociocognitive processes that signal participation played a role in the team dynamics. Combined, these contributes provide advances towards understanding the role that group majority/minority composition has in shaping the underlying foundations in online communication. This understanding has the potential to expand how researchers, practitioners, and educators promote meaningful and engaged communication in diverse teams.

\section{Acknowledgments}
This research was supported in part by The Gates Foundation (INV
- 000752) and The Andrew W. Mellon Foundation (1806-
05902). The authors would like to thank the Next Gen-
eration Undergraduate Success Measurement Project team
members for their efforts in data collection. Thanks to Educational Testing Services for the experiment information.

\bibliographystyle{unsrtnat}
\bibliography{references}  

\end{document}